\documentclass[runningheads]{llncs}
\usepackage[T1]{fontenc}
\usepackage{graphicx}
\usepackage{latexsym}
\usepackage{amssymb}
\usepackage{amsmath}
\usepackage{booktabs}
\usepackage{enumitem}
\usepackage{color}

\usepackage{multirow}
\usepackage{comment}

\begin{document}

\title{Shifted Window Fourier Transform And
Retention For Image Captioning}

\author{Jia Cheng Hu\inst{1}\orcidID{0009-0008-1611-966X} \and
Roberto Cavicchioli\inst{1}\orcidID{0000-0003-0166-0898} \and
Alessandro Capotondi\inst{1}\orcidID{0000-0001-8705-0761}}
\authorrunning{J. C. Hu et al.}
%
\institute{
University of Modena and Reggio Emilia, via G.Campi 213/b, 41125, Modena, Italy \email{name.surname@unimore.it}}

%
%
\maketitle              

\begin{abstract}
Image Captioning is an important Language and Vision task that finds application in a variety of contexts, ranging from healthcare to autonomous vehicles. As many real-world applications rely on devices with limited resources, much effort in the field was put into the development of lighter and faster models. However, much of the current optimizations focus on the Transformer architecture in contrast to the existence of more efficient methods. In this work, we introduce SwiFTeR, an architecture almost entirely based on Fourier Transform and Retention, to tackle the main efficiency bottlenecks of current light image captioning models, being the visual backbone's onerosity, and the decoder's quadratic cost. SwiFTeR is made of only 20M parameters, and requires 3.1 GFLOPs for a single forward pass. Additionally, it showcases superior scalability to the caption length and its small memory requirements enable more images to be processed in parallel, compared to the traditional transformer-based architectures. For instance, it can generate 400 captions in one second. Although, for the time being, the caption quality is lower (110.2 CIDEr-D), most of the decrease is not attributed to the architecture but rather an incomplete training practice which currently leaves much room for improvements. Overall, SwiFTeR points toward a promising direction to new efficient architectural design. The implementation code will be released in the future. 

\keywords{Image-Captioning  \and Retention \and Fourier-Transform}

\end{abstract}

\section{Introduction}
\label{sec:introduction}

Image Captioning 
describes the problem of providing a natural language description of an image without human intervention. It is an important Vision and language task that finds application in a variety of circumstances. For instance, it is adopted in surveillance \cite{albertsson2023analyzing}, 
healthcare monitoring 
\cite{qadeer2023activity}, 
visual question answering \cite{shao2023prompting}, 
autonomous vehicles \cite{dong2023did}, 
and assistance for the visually impaired and blind 
\cite{arystanbekov2023image,ganesan2022deep}. 
However, many applications require the deployment of Image Captioning models in memory and computationally constrained devices which can be problematic. Over the past years, Neural Image Captioning systems achieved outstanding results. Unfortunately, current State-of-the-Art models are often based upon Large Language Models \cite{hu2022scaling,li2022blip}, which are characterized by a massive number of parameters and require abundant memory and computational resources. These aspects hamper their deployment in limited resources devices. 
For this reason, much effort has been put into the design of lighter or more efficient models \cite{lee2021fnet,sun2023retentive,tan2019comic,zhou2022compact}. However, all these architectures are based on the Transformer \cite{vaswani2017attention} which is currently the standard de facto method in Neural Image Captioners defining, for instance, the trend of stateless captioning models \cite{wang2022end,hu2023exploiting}. Recently, the works of \cite{lee2021fnet} and \cite{sun2023retentive,peng2023rwkv} highlighted the limitations of Attention, and the respective authors proposed attractive alternatives such as the Fourier Transform and Retention, characterized by smaller computational cost and number of parameters while performing similarly to the former approach. While their effectiveness has been proven in Language Modelling tasks, to the best of our knowledge, their application in Image Captioning has not been tested yet. In Deep learning, it is hard to predict the behaviour of new methods across different fields. First, when they work well in one field their effectiveness in another can not be implied (Context Shift Problem). Secondly, different methods do not always interact well with each other. As a result, when combined, their final result is of lesser quality when compared to the one produced by the single parts (Synergy Problem \cite{hu2023heterogeneous}).

\begin{figure*}[h]
  \centering
\includegraphics[width=0.9\textwidth]{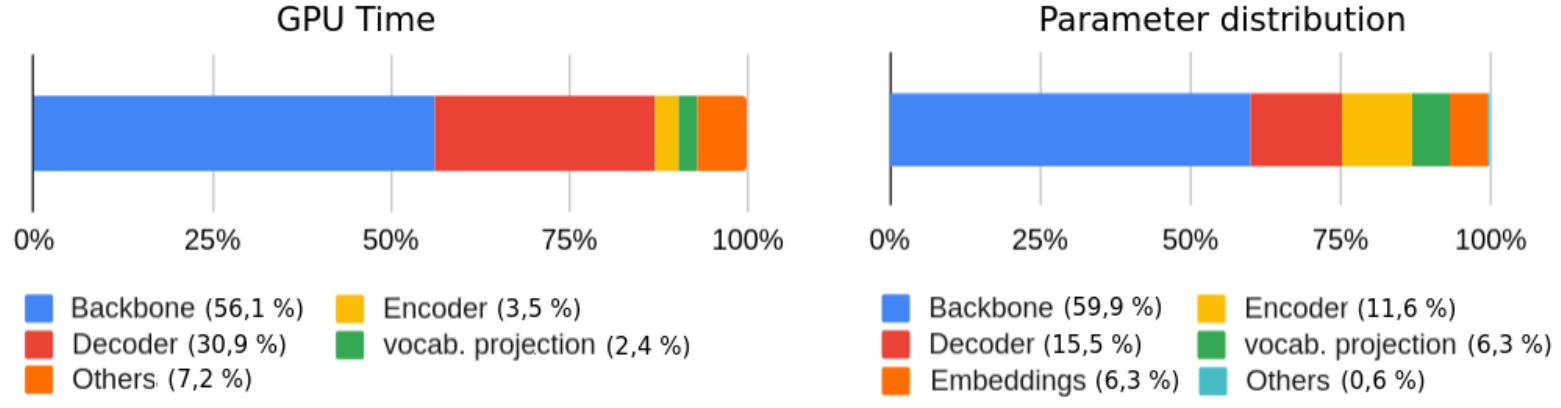}
  \caption{ \label{figure:profiling} Computational and parameters profiling of the baseline Image Captioning model consisting of three layers of Transformer \cite{vaswani2017attention} and Swin-Transformer-S \cite{liu2021swin} backbone. GPU Time is calculated as the average cost to generate captions of images from the COCO \cite{lin2014microsoft} test set.   
  }
\end{figure*}

In this work, we introduce the SwiFTeR architecture which combines the Fourier Transform method in the design of the visual backbone and the Retention in the fusion model decoder. In particular, we highlight the better scaling, higher efficiency, lower memory occupancy and usage compared to State-of-the-Art methods. Additionally, we observe that the decrease in the caption quality is reasonable given the experimental configurations and does not suggest anti-synergy. Overall, the goals and contributions of our work are the following: \emph{(i)} we perform experiments with the recent work of Retention \cite{sun2023retentive} and Fourier Transform \cite{lee2021fnet} in Image Captioning and observe that the two methods enhance model performances at the cost of 4.1 lower CIDEr-D compared to the Transformer model; \emph{(ii)} we present SwiFT, a Shifted Window Fourier Transform visual backbone; \emph{(iii)} Using the previous results, we design SwiFTeR, an architecture characterized by only 20M parameters and capable of generating up to 400 descriptions per-second on NVIDIA GeForce RTX 4090. \emph{(iv)} SwiFTeR currently exhibits a CIDEr-D score of 110 but leaves much room for improvements as it results from a much weaker training setup.

\section{Related Works}
\label{sec:related}


Model compression of Deep Learning models is a hot topic due to the discrepancy between the increasing size of the models and the device these are required to operate on. Because of this, many techniques were developed to decrease the memory and computational cost with the challenge of preserving the quality of the output. Such examples consist of Knowledge Distillation (KD) \cite{hinton2015distilling}, Parameter-Sharing \cite{lan2019albert,hu2024sharebert}, Quantization \cite{zafrir2019q8bert,shen2020q} and Alternative-Formulations \cite{lee2021fnet}. While their effectiveness is often tested on Large Language Models (LLMs) they are less tested in Image Captioning. Our work adopts the KD in the backbone but is orthogonal to other approaches since we focus on the combination of three architectures: the combination of the Swin-Transformer \cite{liu2021swin}, the Fourier Transform layer \cite{lee2021fnet} and the Retention layer \cite{sun2023retentive}. 


Several works tackled the optimization of Image Captioning models or part of it.
Tan et al. \cite{tan2019comic} 
proposed a tiny model based on small Recurrent Neural Networks.
Later, they leveraged sparse structures as well \cite{tan2019image}. 
Wang et al. \cite{wang2023efficient} 
combined the Knowledge Distillation with a lightweight vision backbone and a Small Language Model in the fusion model. 
Tan et al.
\cite{tan2022acort} 
applied embedding token decomposition and parameter-sharing.
Multiple works focused on the Transformer Vision backbone \cite{wang2022towards,nag2023vita,kong2022spvit}, 
Shashank et al.
\cite{nag2023vita} 
proposed optimized operations and pipelines in the inference to increase hardware utilization in edge applications. Kong et al.
\cite{kong2022spvit} 
developed a latency-aware sparsity loss to perform token pruning based on the edge device constraint. Zheng et al.
\cite{zheng2022savit} 
introduced a pruning method that addresses the heterogeneous components inside the vision backbones. In contrast, our work focuses on the architectural side and less on the training strategy or the adoption of additional data to increase the caption quality.

\begin{figure*}[h]
  \centering
  \includegraphics[width=1.0\textwidth]{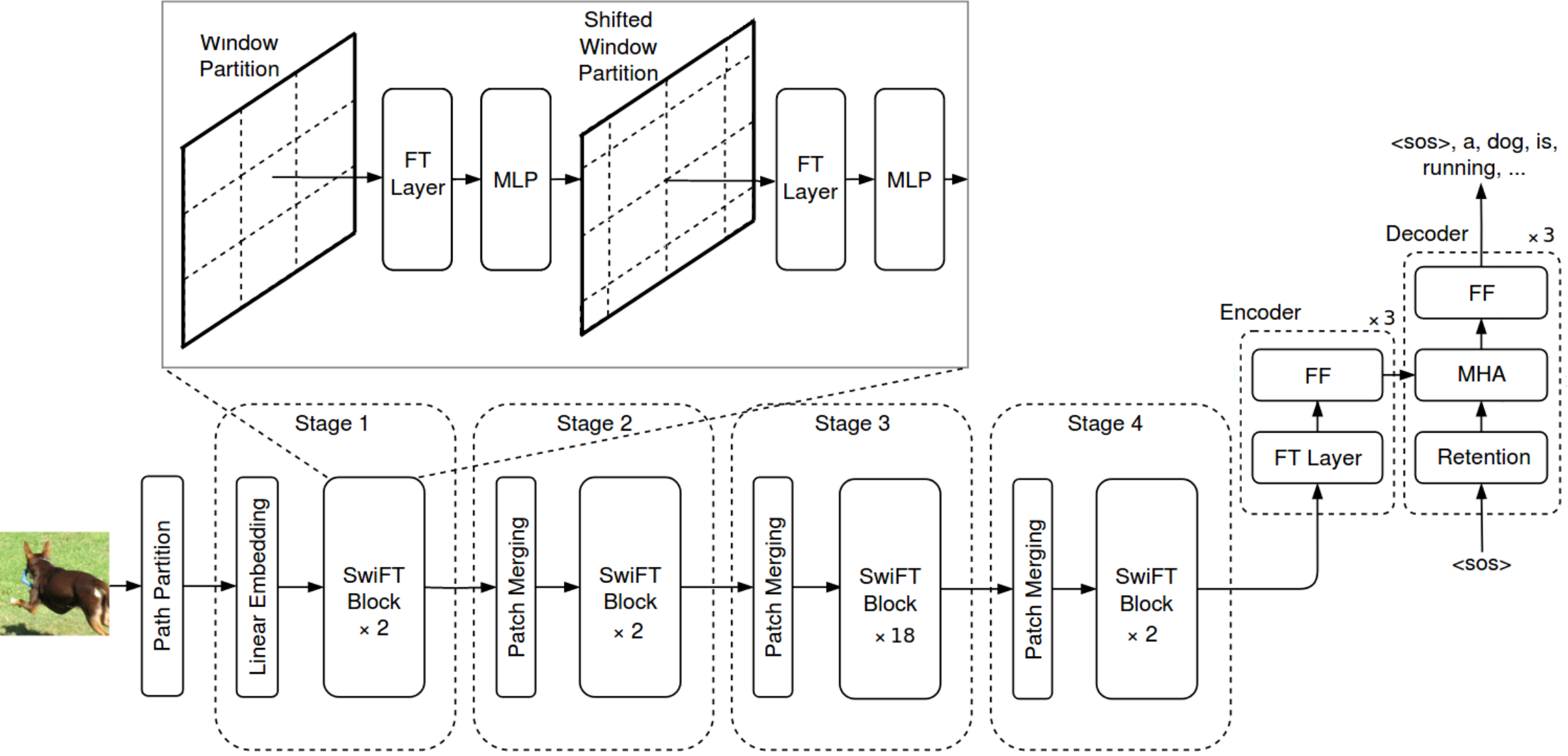}
  \caption{ \label{figure:arch_master_student} The SwiFTeR architecture. Normalization layers and skip connections are omitted.  FT="Fourier Transform", MHA="MultiHead-Attention", FF/MLP="FeedForward".}
\end{figure*}

\section{SwiFTeR Design}
\label{sec:method}

\subsection{Shifted Window Fourier Transform}
\label{sec:shifted_window_fourier_transform}

In Figure \ref{figure:profiling} it can be noted that the backbone's weights and operations occupy the most significant portion of the entire memory and forward cost respectively. To tackle this issue, we propose an adaptation of the Fourier Transform to the Swin-Transformer architecture. 

The Fourier Transform consists of an operation that maps functions defined over the time dimension, into functions from the frequency domain. In \cite{lee2021fnet}, the authors proposed to replace the attention layer with the Discrete Fourier Transform and observed little degradation in the results in exchange for higher efficiency from both computational and memory occupation points of view. Additionally, the operation showcased better computational scalability concerning the number of input elements which is ideal in the case of Image Captioning where the number of visual features can be very large. 

Let $\{\mathbf{x}_0, \mathbf{x}_1, \ldots, \mathbf{x}_{T-1}\}=X\in \mathbb{R}^{T \times H}$ be the input sequence where $\mathbf{x}_{t} \in \mathbb{R}^{H}$ for $t=\{0,\ldots,T-1\}$. The Fourier Transform (FT)  layer introduced in \cite{lee2021fnet}, applies two 1D Discrete Fourier Transform (DFT) on the input. The first, which results are denoted with $\{\mathbf{f}^{seq}_0, \mathbf{f}^{seq}_1, \ldots, \mathbf{f}^{seq}_{T-1}\}=\mathcal{F}^{seq}(X) \in\mathbb{C}^{T\times H}$, is applied over the sequence length and describes the following operations:
\begin{equation}
    \begin{aligned}
    \mathbf{f}^{seq}_k = \sum_{t=0}^{T-1} \mathbf{x}_t e^{-\frac{2 \pi i}{T}tk}, \ \ 0 \leq k \leq T-1 
    \end{aligned}
\label{eq:fnet_dft_seq_len}
\end{equation}

The second DFT, denoted with $\mathcal{F}^{hid}$ shares the same formulation of Equation \ref{eq:fnet_dft_seq_len} but is applied over the hidden dimension. The FT layer is ultimately formulated as:
\begin{equation}
\begin{aligned}
    FT(X) = Real(\mathcal{F}^{hid}(\mathcal{F}^{seq}(X)^{\dagger}))^{\intercal}
\end{aligned}
\end{equation} 
The first DFT mixes the input information along the sequence length and enables the second DFT to distribute the information of the whole sequence across the hidden dimension. The higher-order modules, e.g. the Feed-Forward, can now process the input in a bidirectional and since FT is a parameter-less operator, they hold the responsibility of learning to form meaningful compositions of the input.

Given an image or activation map of of size $A=(H, W, C)$ where $H$ is the height, $W$ is the weight and $C$ is the number of channels, the Windowed Fourier Transform (WFT) partitions the input into several square and rectangular windows and applies the Fourier Transform. By omitting the shifted mechanism and border cases (we refer to \cite{liu2021swin} for details), in its essence, the layer partitions the input into $N$ windows of size $A'=(H', W', C')$, then it treats each partition as a sequence of size $\mathbb{R}^{H'W' \times C'}$ and applies:
\begin{equation}
\begin{aligned}
    WFT(A') = Real(\mathcal{F}^{hid}(\mathcal{F}^{seq}(A')^{\dagger}))^{\intercal}
\end{aligned}
\label{eq:swin_ft}
\end{equation}

The advantage of the FT layer is two-fold. From the perspective of memory occupation, it does not require learned parameters. Second, the DFT can be computed using the Fast Fourier Transform Algorithm, 
which exhibits a complexity of O($T log T)$ compared to the O($T^2$) of the Self-Attention \cite{vaswani2017attention}.

\subsection{Retention Layer}

RetNet \cite{sun2023retentive} describes a novel network where the key idea consists of integrating a recurrent aspect in the Key-Value-Query approach proposed by \cite{vaswani2017attention}. While the latter approach accelerates the training phase by enabling parallelism, one of the main drawbacks of RNNs, the Transformer, during the inference stage is slower. This happens because during each time step $t$, the decoder is required to process the entirety of previous $t$ input elements, which is much slower compared to the single hidden state propagated through time. 

To address this issue, \cite{sun2023retentive} introduces the Retention Layer, which proposes a formulation that can be computed either recurrently or in a parallel manner. Let $\{\mathbf{x}_1, \mathbf{x}_2, \ldots, \mathbf{x}_T\} = X \in \mathbb{R}^{T \times H}$ be the input sequence of vectors where $\mathbf{x}_t \in \mathbb{R}^{H}$ for $t=\{1,\ldots,T\}$. Assuming for simplicity, the Single-Scale formulation and omitting the final output gate, first, the Retention computes $Q, K, V \in \mathbb{R}^{T \times H}$ as:
\begin{equation}
\begin{aligned}
    Q = (XW_Q) \odot \Theta, \ \ \ K = (XW_K) \odot \overline{\Theta}, \ \ \ V = XW_V
\end{aligned}
\end{equation}
where $W_Q, W_K, W_V \in \mathbb{R}^{H \times H}$ are linear mapping. Ignoring for the moment $\Theta$, $\overline{\Theta}$ and $D$ it can be proven that the following parallelizable equation:
\begin{equation}
    Retention_{PLL}(X) = (Q K^{\intercal} \odot D)V
\label{eq:parallel_retention}
\end{equation}
is equivalent to the recurrent formulation:
\begin{equation}
\begin{aligned}
    S_t = \gamma S_{t-1} + K_t^{\intercal}V_t  \ \ \ \ \ \ 1 \leq t \leq T \\ 
    Retention_{RNN}(\mathbf{x}_{t}) = Q_t S_t  \ \ \ \ \ \ 1 \leq t \leq T 
\end{aligned}
\label{eq:recurrent_retention}
\end{equation}
where $Q_t, S_t, x_t \in \mathbb{R}^{H}$ and $S_t$ is the state vector. 

The equivalence between Equation \ref{eq:parallel_retention} and Equation \ref{eq:recurrent_retention} is 
made possible thanks to the definition of $\Theta$ and $\overline{\Theta}$, $D$ and $\gamma$. While we refer to \cite{sun2023retentive} for the proof, in this Section we propose some insights. \emph{(i) Causality.} A causal mask, denoted with $D \in \mathbb{R}^{T \times T}$ in Equation \ref{eq:parallel_retention} a causal mask is multiplied to $Q$ and $K$. This prohibits the parallel formulation from looking at the future elements, in the absence of such a condition, a recurrent formula would not be feasible. \emph{(ii) Linearity.} Only linear operations are involved. Therefore, the equivalence can be expressed by the following equation $Q_t S_t$ = $\sum_{m=0}^{t} Q_t A^{t-m} K_{m}^{\intercal}V_{m}$ where $S_t$ = $AS_{t-1} + K_t^{\intercal}V_t$ and $A \in \mathbb{R}^{H \times H}$. which leads to the last insight. 
\emph{(iii) Computational feasibility.} While $A$ is absorbed into $W_{K}$ and $W_{Q}$, the elements $\gamma \in \mathbb{R}^{H}$, and $\Theta, \overline{\Theta} \in \mathbb{R}^{T \times H}$ are defined and chosen such that $A^{t-m}$ can be decomposed into a numerically more stable and easier to compute formulation. 

Retention is adopted in the fusion model decoder for multiple reasons. First, in \cite{sun2023retentive} they showcased that Retention achieves comparable performances with Attention-based models which suggest its effectiveness in understanding image descriptions. Second, the recurrent formulation directly addresses the performance bottleneck of the decoding stage, highlighted in Figure \ref{figure:profiling}. In particular, assuming the cost of a single decoder iteration is quantified by the number of vectors processed at each time step, it can be reduced to O(1) thanks to the recurrent formulation, compared to O($T$) in the case of Transformer. If we then consider a final image description of length $T$, the cost of Greedy Search for the recurrent module is O($T$), in contrast to $O(1) + O(2) + \ldots + O(T) = O(T^{2})$ of the stateless decoder. Finally, thanks to the dual formulation, the layer can also be trained efficiently, which can be a desirable property in the case of online training where updates need to be made frequently.


\begin{table*}[h]
\centering
\footnotesize
\caption{\label{tab:ablation} 
Ablation study of SwiFTeR architecture. The performance is evaluated on the COCO validation dataset. Top) reports the "fusion model only" ablation study. The Swin-Transf.-S backbone is adopted but the backbone parameters, FLOPs and latency are excluded from the calculation. Captions for quality measuring are generated with beam search with a beam size of 3 in the case of Cross-entropy (XE) training and a beam size of 5 in the case of reinforcement (RF) training. Greedy decoding is adopted for the latency computation. Bottom) reports the values for backbone+fusion model.  
The latency is measured for a single forward pass, assuming 49 visual features and 20 tokens. $\gamma$ denotes the latency percentage compared to the baseline value. 
Latency is computed on a single Mobile NVIDIA GeForce RTX 3070 Ti GPU. "$\star$"=Baseline.
}
 \begin{tabular}{| l | c | c | c c c |}

\hline
Fusion model only & Train & B4 \ \ \ \   C \ \ & Params & FLOP & Latency ($\gamma$) \\
\hline
Normal & {} & {} & {} & {} & {} \\
\ \ \ \ 
Transf. Enc. \& Transf. Dec. ($\star$) & XE & 34.3  \ 114.5 & 32.7M & 3.7G & 37.0ms (100\%) \\
\ \ \ \ Tranfs. Enc. \& RetNet Dec. & XE & 33.6 \ 113.2  & 35.8M & 4.1G & 28.7ms (77.5\%) \\ 
\ \ \ \  RetNet Enc. \& Transf. Dec. & XE & 32.2 \ 107.5  & 35.8M & 6.5G & 36.0ms (97.0\%) \\
\ \ \ \  RetNet Enc. \& RetNet Dec. & XE & 32.1 \ 106.7  & 39.0M & 6.9G & 31.2ms (84.3\%) \\
\ \ \ \ FNet Enc. \& Transf. Dec. & XE & 33.4 \ 112.4 & 29.6M & 2.7G & 35.6ms (96.2\%) \\
\ \ \ \ FNet Enc. \& RetNet Dec. & XE &  32.8 \ 110.4  & 32.7M  & 3.1G & 28.9ms (78.1\%) \\
Small & {} & {} & {} & {} & {} \\
\ \ \ \ FNet Enc. \& RetNet Dec. & XE & 34.3 \ 111.5  & 2.7M & 0.1G & 28.0ms (75.6\%) \\
\hline
\multicolumn{6}{c}{} \\
\hline
Backbone included & Train &   B4 \ \ \ \  C \ \ & Params & FLOP & Latency ($\gamma$) \\
\hline
Normal (Swin-Transf.-L) & {} & {} & {} & {} & {} \\
\ \ \ \ FNet Enc. \& RetNet Dec. ($\star$) & XE & 35.3 \ 118.9 & 229.7M & 107.0G & 55.4ms (100\%) \\
\ \ \ \  FNet Enc. \& RetNet Dec. & RF &  39.1 \ 134.2 & 229.7M & 107.0G & 55.4ms (100\%) \\ 
Small (Swin-Transf.-S) & {} & {} & {} & {} & {} \\
\ \ \ \ FNet Enc. \&           
RetNet Dec. & XE & 34.0 \ 111.4 & 52.7M & 8.8G & 49.0ms (88.4\%) \\ 
\ \ \ \ FNet Enc. \& RetNet Dec. & RF & 36.8 \ 124.8 & 52.7M & 8.8G & 49.0ms (88.4\%) \\
SwiFTeR (SwiFT$_{weak}$) (Ours) & {} & {} & {} & {} & {} \\
\ \ \ \ FNet Enc. \& RetNet Dec. & XE & 30.5 \ \ 94.7 & 20M & 3.1G & 46.1ms (83.2\%) \\
\ \ \ \    FNet Enc. \& RetNet Dec. & RF & 33.0 \ 109.5 & 20M & 3.1G & 46.1ms (83.2\%) \\

\hline
\multicolumn{6}{l}{Normal=\{H=512, F=2048\}, Small=\{H=96, F=384\}} \\
\multicolumn{6}{l}{H=Hidden dimension, F=Feed forward size.} \\

\end{tabular}
\end{table*}


\subsection{SwiFTeR}

SwiFTeR stands for Swin-Fourier Transform and Retention Network. It comprises the SwiFT$_{weak}$ backbone, which is made by replacing the Attention with the operation described in \ref{sec:shifted_window_fourier_transform} in the Swin-Transf.-S. And a fusion model, that consists of FNet layers within the encoder and Retention in the decoder.
Formally, the input image $I \in \mathbb{R}^{C \times H \times W}$
is fed to the SwiFT$_{weak}$ backbone which generates the initial set of processed visual features  $X_{0}=\{x^{0}_1$, $x^{0}_2$$,\ldots,$$ x^{0}_{|X|}\}, \ x^{0}_{i} \in \mathbb{R}^{d_m}$:
\begin{equation}
    \begin{aligned}
        X_{0} &= SwiFT_{weak}(I) 
    \end{aligned}
\end{equation}
A linear mapping $W_{I} \in \mathbb{R}^{d_m \times H}$ is applied to conform the visual features to the hidden dimension of the fusion model. The fusion model consists of $N$ encoders and $N$ decoders. Let us assume for simplicity $X_0 \in \mathbb{R}^{H}$, the $n$-th  encoder is defined as:
\begin{equation}
    \begin{aligned}
        E_{n} &= X_{n-1} + FT(Norm^{FT}_{n}(X_{n-1})) \\
        X_{n} &= E_{n} + FF_{n}(Norm^{FF}_{n}(E_{n}))
    \end{aligned}
\end{equation}
whereas, the $n$-th decoder is defined as:
\begin{equation}
    \begin{aligned}
        D_{n} &= Y_{n-1} + Retention_{n}(Norm^{D}_{n}(Y_{n-1})) \\
        W_{n} &= B_{n} + Attention_{n}(Norm^{CA}_{n}(D_{n}), X_{N_{enc}}) \\
        Y_{n} &= W_{n} + FF_{n}(Norm^{FF}_{n}(W_{n}))
    \end{aligned}
\end{equation}
where $Y_{0}=\{y^{0}_1, y^{0}_2, \ldots, y^{0}_{|Y|}\}, \ y^{0}_{i} \in \mathbb{R}^{H}$.
All layers are summed through a linear projection and the final output is fed to the classification layer. The complete architecture is depicted in
Fig. \ref{figure:arch_master_student}.


\section{Experimental Setup}

\subsection{Dataset and Training}

The visual backbone, SwiFT$_{weak}$, is trained on ImageNet-1k \cite{deng2009imagenet}, which is made of only million images belonging to 1000 classes (reason why is a \textit{weak} backbone). Images are resized into $3\times244\times244$ and each channel is normalized to mean=$(0.485, 0.456, 0.406)$ and standard deviation of $(0.229, 0.224, 0.225)$. The training of the fusion model and evaluation of the whole architecture is conducted on the renowned Microsoft COCO benchmark \cite{lin2014microsoft}, which consists of 113 thousand images for training, 5000 for the validation and 5000 for the evaluation test set. Each image is paired with five descriptions. Each reference caption is pre-processed by lowering casing and punctuation filtering. The vocabulary is made of tokens that occur at least five times resulting in 10000 unique words.

\begin{table*}[t]
  \centering
  \small
\caption{ \label{tab:sota_evaluation}
Offline comparison of state-of-the-art models over the COCO Karpathy test split. "Pre-training" in the column "Type" denotes the fusion model size and the practice of pre-training the final model (on top of the already pre-trained backbones) on additional text or images before fine-tuning the COCO test split. In "Caption quality" the subscript $XE$ and $RF$ distinguish between the results with Cross-Entropy and Reinforcement training. In conformity with the literature, the FLOPs computation between the image backbone and fusion model differs. Fusion model FLOPs include multiplications and sums and we assume a length of 128 whereas image backbone FLOP computation does not count for additions.}

\begin{tabular}{ | c | c | c | c  c  c | c | c |}
 \hline
\multicolumn{2}{|c|}{Model} & \multicolumn{1}{c|}{Backbone} & \multicolumn{3}{c|}{Caption quality} & \multicolumn{2}{c|}{Performance} \\
\hline
Type & Architecture & Image & C$_{XE}$ & B4$_{RF}$ & C$_{RF}$ & Params & FLOP \\
\hline
\multirow{4}*{ 1
} 
 & Att2all \cite{rennie2017self} & R-CNN & 99.4 & 34.2  & 114.0 & 60M & >760G \\
{} & UpDown \cite{anderson2018bottom} & F-RCNN$_{101}$ & 113.5 & 36.3 & 120.1 & 69M & 762G \\
{} & AoANet \cite{huang2019attention} & F-RCNN$_{101}$ & 119.8 & 38.9 & 129.8 & 90.7M & 766G \\

{} & ExpansionNet v2 \cite{hu2023exploiting} & Swin-Transf.-L  & 128.7 & 41.5 & 140.4 & 245M & 115.3G \\
\hline
\multirow{3}*{
2
} & UNIMO$_{B}$ \cite{li2020unimo} & F-RCNN$_{101}$ & 124.4 & - & - & 172.8M & 789.5G \\
{} & LEMON$_{B}$ \cite{hu2022scaling} & ResNeXt$_{152}$ & 133.3 & 41.6 & 142.7 & 250.2M & 1029.7G \\
{} & BLIP$_{B}$ \cite{li2022blip} & ViT$_B$ & 133.3 & - & - & 195.4M & 78G \\
\hline
\multirow{4}*{
3
} & E2E-VLP \cite{xu-etal-2021-e2e} & ResNet-152 & 117.3 & - & - & 94M & 12.8G \\
{} & MiniVLM \cite{wang2020minivlm} & Eff-DET & 119.8 & - & - & 33M & 8.3G \\
{} & DistilVLM \cite{fang2021compressing} & Eff-DET & 120.8 & - & - & 33M & 8.3G \\
{} & LightCap \cite{wang2023efficient} & ResNet-50 & 125.8 & 40.1 & 136.6 & 38M & 5.3G \\
\hline
\multirow{2}*{
4
} & Ablation  & Swin-Transf.-S & 113.2 & 36.5 & 125.5 & 52.7M & 8.8G \\
{} & SwiFTeR & SwiFT$_{weak}$  & 96.0 & 33.3 & 110.2 & 20M & 3.1G \\
\hline
 \end{tabular}
\end{table*}

We train the backbone with Knowledge Distillation (KD) \cite{hinton2015distilling} using the Swin-Transformer-S (Small) \cite{liu2021swin} as the teacher. The student model SwiFT$_{weak}$, is trained on the ImageNet-1k \cite{deng2009imagenet} classification task and compute the loss as the KL-divergence on the output prediction layers summed to the MSE of the last feature maps. No image augmentation is adopted. The KD is adopted to accelerate the training rather than increasing the performances, due to the heterogenity between the teacher and student architectures.


Regarding the Fusion Model, we follow the standard practice of training models on Cross-Entropy Loss for Image Captioning \cite{hu2023exploiting}, followed by CIDEr-D optimization \cite{rennie2017self}, described by the following minimization:
\begin{equation} 
    min_{\theta} \ \ L_{SCST}(\theta)=-\mathbb{E}_{p_{\theta}}[r(y^{s}_{1:T})]
\end{equation}
where $Y^{s}_{1:T} = (y_1, y_2, \ldots, y_T)$ is the sampled description and $r$ is the CIDEr-D reward function. 
We adopt the Standard Self-Critical Sequence Training \cite{Hu202339}\footnote{Signature: STANDARD\_wInit+Cider-D[n4,s6.0]+average[nspi5]+1.0.0}  and follow the training strategy of \cite{hu2023exploiting} up until the third step.

\section{Results}

\subsection{Ablation Study}

In the ablation study, we compare the performance and costs of our proposed architectures with several model variants. Since SwiFTeR changes both the backbone and the fusion block of the baseline end-to-end Transformer architecture, we split the discussion into two paragraphs. The first discusses the changes in the fusion network only. The latter analyzes the impact of changing the backbone.


In Table \ref{tab:ablation}-Top we report the first ablation study which focuses on the cross-modal fusion network and uses the Base Transformer as a baseline.  In detail, we keep the backbone of the Swin-Transformer-S and observe the impact of changing the encoder and decoder architectures. It can be noted that Retention in the decoder leads to a negligible performance degradation of 1.3 CIDEr-D, whereas, a significant drop of 7 CIDEr-D can be observed when adopted in the encoder. While the first degradation is acceptable as a trade-off with a much more scalable inference time, the latter is caused by the formulation 
of the retention which introduces intrinsic sequentiality bias and is not helpful in the visual processing. Instead, adopting the Fourier Transform in the encoder leads to an increase of 2.9 CIDEr-D compared to the Retention case and a decrease of only 1.9 CIDEr compared to the self-attentive encoder. Overall, the proposed fusion network of FNet encoder \&  RetNet decoder leads to a degradation of 4.1 CIDEr-D, however, we adopt this configuration in SwiFTeR as it compensated by a decrease in terms of FLOPs and latency as reported in Table \ref{tab:ablation}-Top. Notably, the metrics of FLOPs and latency hide the true advantages of adopting the Retention layer. For instance, Retention exhibits higher FLOPs but latency is smaller. Additionally, the superior scalability of retentive networks to long sequence lengths is understated in Table \ref{tab:ablation}. This aspect will be clarified more in Section \ref{section:latency_comparison}.

Overall, we chose the configuration of "FNet Enc. and RetNet Dec." as the fusion model for SwiFTeR for efficiency reasons and because the combination of the two strategies does not showcase a significant quality degradation, therefore anti-synergy. However, it does not provide a noticeable reduction in memory occupancy. This aspect can be addressed by decreasing the model size. In the typical configuration, models have three layers and a hidden dimension of at least $H$=512 \cite{huang2019attention,hu2023exploiting}. Instead, we set the hidden dimension to $H$=96 and label this version with "Small" in Table \ref{tab:ablation}-Top. This configuration is not only lighter but, surprisingly, it achieves slightly better CIDEr compared to the larger model, which suggests a high degree of overfitting in the previous setup. 


One of the most significant costs of a Captioning system, as depicted in Figure \ref{figure:profiling} is given by the backbone. In Table \ref{tab:ablation}-Bottom we report the impact of using different backbones on the performances. As expected, the Swin-Transformer-L (Large) backbone leads to the highest caption quality. When the fusion model is tested on Swin-Transformer-S, it achieves about 10 lower CIDEr-D score but is $\times4.35$ smaller and $\times12.15$ less computationally expensive. Finally, replacing the Swin-Transformer-S backbone with SwiFT$_{weak}$, we further reduce the parameters, achieving a $\times11.48$ compression, and $\times34.51$ reduction in FLOPs. However, a significant decrease in caption quality is observed due to a weak training configuration, which makes SwiFT a much weaker visual backbone. This resulted from adopting ImageNet-1k instead of ImageNet-22k in contrast to the Swin-Transformer-S. Additionally, no data augmentation was performed, which is key to achieving a robust backbone \cite{liu2021swin}. These aspects will be addressed in future works but the weak training configuration does not affect the memory requirements and efficiency, the focus of our work.

\subsection{Comparison With State-of-the-Art}

In Table \ref{tab:sota_evaluation} we existing categorized State-of-the-Art models into three classes: Type-1:"Normal without Pre-train", Type-2:"Large with Pre-train", and Type-3:"Light with Pre-train".
Type-1 models \cite{hu2023exploiting,wang2022end,anderson2018bottom}, refer to models that consist of a strong image backbone and a standard sized fusion model trained from scratch on the MS-COCO dataset only, without additional training data. Whereas, Type-2 models \cite{hu2022scaling,li2022blip,li2020unimo},
are characterized by a strong image backbone and a large fusion model which is further pre-trained on a large amount of additional data to build the foundation model for multiple fine-tuning tasks, image captioning being one of them. Both categories focus on achieving high caption quality regardless of the computational and memory cost. As a result, while they achieve up to 142.7 CIDEr-D, they are associated with hundreds of GFLOPs and hundreds of millions of parameters. 
In contrast, the Type-3 category \cite{xu-etal-2021-e2e,wang2020minivlm,fang2021compressing,wang2023efficient} focus on efficiency but also preserving a high description quality. To achieve that, they adopt a lightweight image backbone and a small fusion model architecture that is first initialized with language modeling weights and then pre-trained on a large amount of additional image-text data. The most remarkable example is LightCap \cite{wang2023efficient} which exhibits 136.6 CIDEr-D on the MS-COCO test set, using only 38M parameters and up to 5.3 GFLOPs for a single image, which is the State-of-the-Art architecture among efficient models.
Our proposal falls into a fourth model category, Type-4:"Light without Pre-train". SwiFTeR aims for high inference speed and model size reduction. It is made of 20M parameters and requires up to 3.1 GFLOPs, which is up to $\times$12.5 smaller and requires $\times$331 less operations compared to large models. It is 18M parameters smaller ancad requires $\times$1.7 less operations than LightCap. In terms of caption quality, SwiFTeR exhibits 110.2 CIDEr-D, which is similar to Att2al \cite{rennie2017self}, E2E-VLP \cite{xu-etal-2021-e2e}, MiniMLM\cite{wang2020minivlm} and DistilVLM \cite{fang2021compressing}, while being much faster and smaller. On the downside, SwiFTeR is currently less descriptive than LightCap by 26.4 CIDEr-D. However, this is attributed to a much weak training configuration in the backbone SwiFT$_{weak}$ compared to LightCap. 
In the ideal training configuration (backbone trained on ImageNet-22k), Retention and Fourier Transform achieve the expected results of 125.5 (see "SwiFTeR ablated" in Table \ref{tab:sota_evaluation}). To further support this claim, we note that the decrease is not to be attributed to the lack of parameters since EffcientVLM \cite{fei2022efficient} achieves 120.8 using a backbone of a similar number of parameters. Therefore, the current results leave much room for improvement. 

\subsection{Efficiency}
\label{section:latency_comparison}

Despite the Retention layer contributing to a small fraction of the architecture, it plays a key role in decreasing the overall latency of the captioning model. In Table \ref{tab:sota_evaluation} we reported the FLOPs assuming an output sequence length of $128$, to be processed in a single passage. However, in reality, captions are generated step by step and the real computation is higher than what is typically reported for evaluation in literature. Additionally, FLOPs do not reflect the time cost which depends instead on synchronizations and parallel operations. For these reasons, the recurrent formulation of SwiFTeR, enabled by the Retention layer, cannot be fully appreciated in the previously discussed comparisons.

\begin{figure}[h]
  \centering
\includegraphics[width=1.0\textwidth]{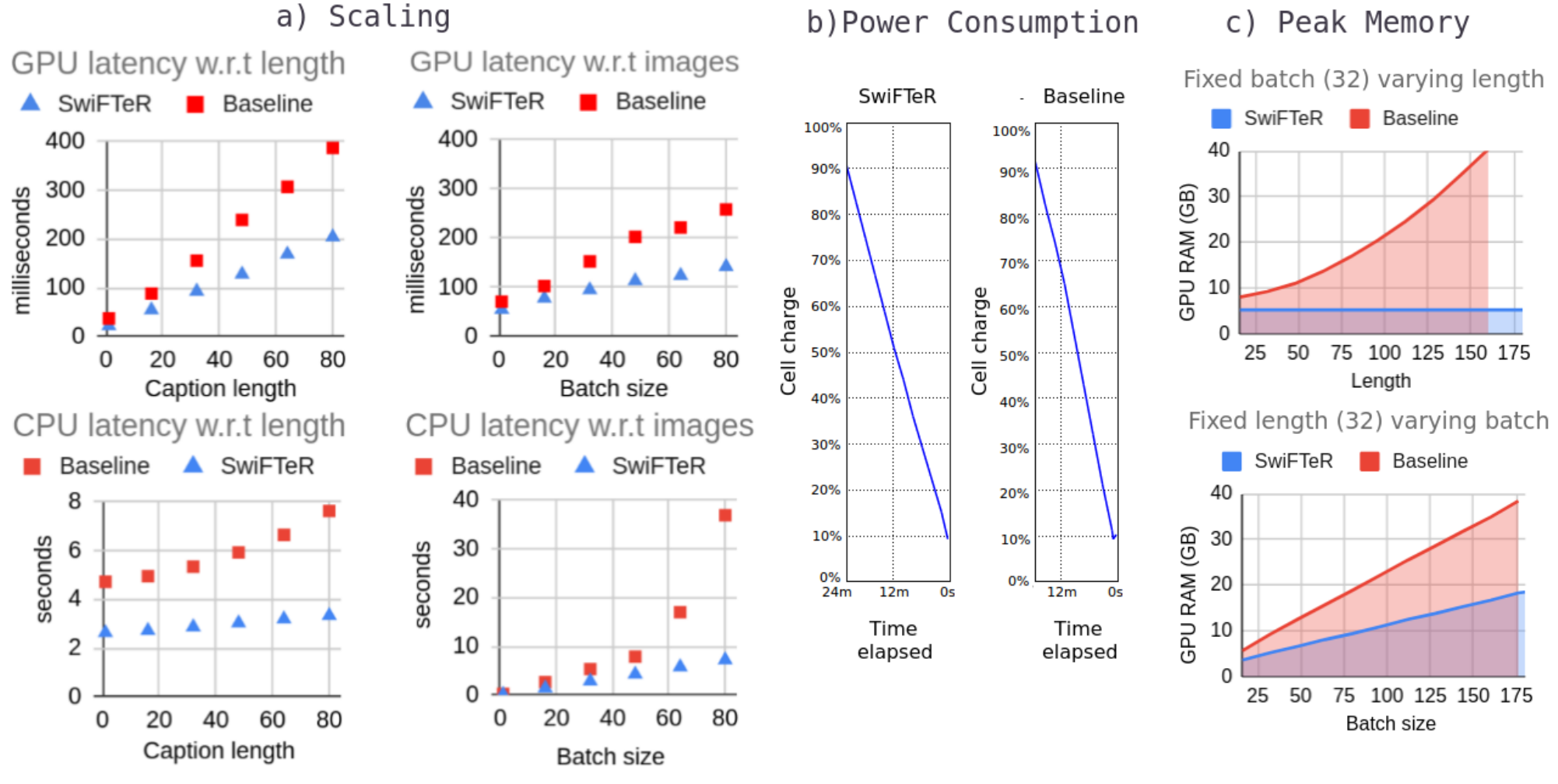}
  \caption{ \label{figure:scaling} Latency of SwiFTeR and baseline models with different length and batch size configurations. Unless differently stated, the length is set to 32, and the batch size is set to 32.
  The GPU consists of an NVIDIA GeForce RTX 4090 24GB. Each configuration is accommodated into one batch. Intel(R) Core i9-13900K is adopted as CPU. 
  }
\end{figure}

To address this issue, we create a baseline architecture consisting of the Swin-Transf.-S + Transformer fusion model (the baseline) with the same hidden dimension as SwiFTeR's fusion model (H=96). Such a baseline well represents the behaviour of SotA models in Table \ref{tab:sota_evaluation} since they are based upon the same architecture. 

Figure \ref{figure:scaling} highlights the time costs and several observations can be made from it. By looking at the GPU times (Figure \ref{figure:scaling}-Top Row), it can be seen that SwiFTeR scales better with increased caption length and batch size. However, both latency plots appear to vary linearly to the input change. This behaviour is due to the selected GPU, the NVIDIA GeForce RTX 4090 24GB, a high-performing Desktop GPU. Because of the low computational requirements of both SwiFTeR and the baseline, most of the time is spent on memory transfer and synchronization, hence the linearity. The GPU was 
selected for its capacity to store all the experimental configurations in RAM, but it is not a good representative of limited resources devices, such as IoT and embedded systems. Instead, CPU times (Figure \ref{figure:scaling}-Bottom Row) reflect the exact number of operations required by the two architectures in each configuration. Hence, it provides a better idea of their efficiency on devices with limited computational capabilities. The time cost on the CPU showcases the benefit of both the Retention-based fusion model and the SwiFT backbone. When the caption length increases, the final cost is dominated by the cost of the fusion decoder. As a result, the first design choice translates into a linear time increase since the cost of handling a recurrent state is constant. In contrast, the baseline architecture exhibits a polynomial increase with respect to the caption length (Figure \ref{figure:scaling}-Bottom-Left). When the batch size or the number of images increases, the cost is dominated by the cost of the backbone. In this case, SwiFTeR latency showcases a linear behaviour as expected. However, the baseline's latency is not only significantly higher, but the high peak memory requirements led by the baseline, forced the RAM into paging, which explains the quadratic behaviour.

Overall, running SwiFTeR on unoptimized PyTorch code\footnote{PyTorch is not optimized for efficiency. Latency can be reduced significantly using TensorRT and CUDA implementations.} requires up to 180 ms on Intel(R) Core i9-13900K CPU, and about 54 ms on NVIDIA GeForce RTX 4090 GPU, to process a single image of size 224$\times$244 and produce a caption of length 32, which is comparable to State-of-the-Art proposals such as DistilVLM \cite{fang2021compressing}, MiniVLM \cite{wang2020minivlm}, and LightCap \cite{wang2023efficient}. However, it showcases scalability to caption length and the smaller memory enables more images to be processed simultaneously, making it suitable for applications where many executions are needed frequently.

\subsection{Power Consumption and Memory}

While FLOPs in Table \ref{tab:sota_evaluation} do not reflect well the latency due to parallelism and synchronization aspects, it can be indicative of how much power is consumed to run the model. To test this aspect, we execute the Baseline architecture defined in Section \ref{section:latency_comparison} and SwiFTeR on Asus VivoBook equipped with NVIDIA GTX GeForce 1050 and Intel Core i7-7700HQ. In Figure \ref{figure:scaling}-b) it appears that generating on GPU, descriptions of length 32 from batches of 8 images at 10Hz, the Baseline discharges the cell from 90\% to 10\% in 14 minutes and 37 seconds, which means it is $\times1.5$ more power-consuming than the case of SwiFTeR, which does so in 22 minutes and 7 seconds. 

Another advantage of our proposal lies in the peak memory usage, which can be seen in Figure \ref{figure:scaling}-c). SwiFTeR keeps the memory peak constant regardless of the caption length. Whereas, the memory slope is lower in case of increased batch size, which enables more images to be processed simultaneously.


\section{Conclusion and Future Works}

In this work, we proposed a high-efficiency and lightweight Image Captioning model named SwiFTeR, characterized by Shifted Window Fourier Transform as the visual backbone and Retention in the fusion model. We assessed that Retention and Fourier Transform work well in the context of Image Captioning beyond the Language Modelling tasks they were originally designed for. Experimental results on the fusion model only showcased that the two methods led to negligible degradation in caption quality compared to the Transformer. Whereas, experimental results on the entire SwiFTeR architecture highlighted greater efficiency, compared to established architectures, thanks to superior scalability to the caption length and the image size. These aspects in addition to the lower power consumption, lower peak RAM usage, and small storage memory requirements make our proposal particularly suitable for resource-limited devices. Since the present paper focused mostly on designing a lightweight and fast architecture, future works will address the caption quality aspect by designing a stronger visual backbone and pre-training the fusion model on additional training data, following the practice of the most established works \cite{wang2023efficient,fei2022efficient,li2022blip,li2020unimo,hu2022scaling}.

%
%
%
\bibliographystyle{splncs04}
\bibliography{mybibfile}

\end{document}